%% file: main.tex
\pdfoutput=1  

\PassOptionsToPackage{dvipsnames}{xcolor}

\documentclass{article}
\usepackage{iclr2027_conference,times}

\newif\ifmoveoverflow
\moveoverflowtrue

\input{preamble}

\usepackage[pagebackref=false,breaklinks,colorlinks,allcolors=iclrblue]{hyperref}
\usepackage{url}

\usepackage[capitalize]{cleveref}

\title{Timo: \textbf{T}aming Mult\textbf{i}modal Diffusion Transformers for Human \textbf{Mo}tion Generation}

\author{\textbf{Zhao Wang, Jiangtao Hu, Jack Yu, Tao Yu\textsuperscript{\dag*}}\\
LimX Dynamics\\
{\small \textsuperscript{\dag}Project Lead \quad
 \textsuperscript{*}Corresponding Author}
}

\iclrfinalcopy 

\begin{document}

\maketitle

\input{sec/0_abstract}

\input{sec/mov/fig_radar}

\input{sec/1_intro}
\input{sec/2_related}
\input{sec/4_method}

\input{sec/3_data}

\input{sec/5_experiments}

\input{sec/6_conclusion}

\input{sec/7_statements}

\bibliographystyle{iclr2027_conference}
\bibliography{main}

\input{sec/X_suppl}

\end{document}

%% file: preamble.tex
\usepackage{xspace}
\usepackage{xcolor}   
\usepackage{graphicx}
\usepackage{amsmath}
\usepackage{amssymb}
\usepackage{booktabs}
\usepackage{multirow}
\usepackage[shortlabels,inline]{enumitem}

\usepackage[format=plain,labelformat=simple,labelsep=period,font=small,%
            compatibility=false]{caption}
\usepackage[font=footnotesize,skip=3pt,subrefformat=parens]{subcaption}
\usepackage{float}
\usepackage{placeins}
\usepackage{needspace}

\definecolor{iclrblue}{rgb}{0.21,0.49,0.74}

\usepackage{microtype}

\newcommand{\method}{Timo\xspace}



%% file: sec/0_abstract.tex
\begin{abstract}
Most existing human motion generation (HMG) methods use cross-attention
modules to inject text semantics, but ignore the importance of bidirectional
modeling between motion and text tokens, which limits text comprehension.
A straightforward idea is introducing multimodal diffusion transformers
(MMDiT), which have shown effective joint text--visual modeling in vision
generation, into HMG. 
However, we find that articulated motion is temporally coherent but weakly correlated
across joints, in which directly applying an MMDiT with flow matching produces poorly
coordinated and jerky motion.
In this work, we propose \method{}, a novel kinematics-aware MMDiT framework
tailored for HMG. \method{} combines fully shared multimodal attention for
bidirectional text--motion modeling with flow matching, geometric and
rotational-kinematics supervision that compares actual rotations and their
changes over time, and a two-stage curriculum
progressing from broad motion learning to detailed caption alignment.
Further, we construct a benchmark of $40{,}025$ held-out clips from six public
datasets spanning diverse actions, assessing six complementary dimensions
under a common evaluator and scoring protocol. Our model substantially
outperforms state-of-the-art methods in both quantitative and qualitative
evaluations. Remarkably, \method{} surpasses Kimodo on five of six dimensions,
achieving a $40.8\%$ relative improvement in the average benchmark score.
Project page: \url{https://kyfafyd.wang/projects/timo}. Demo page: \url{https://timo.kyfafyd.wang}.
\end{abstract}

%% file: sec/mov/fig_radar.tex
\begin{figure}[H]
  \centering
  \includegraphics[width=0.59\linewidth]{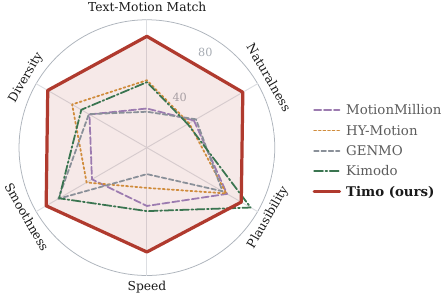}
  \caption{\textbf{Comparison on our six-dataset benchmark.}
  \method{} and four released systems: MotionMillion~\citep{fan2025motionmillion},
  HY-Motion~\citep{wen2025hymotion}, GENMO~\citep{genmo2025} and
  Kimodo~\citep{rempe2026kimodo}.}
  \label{fig:radar}
\end{figure}

%% file: sec/1_intro.tex
\section{Introduction}
\label{sec:intro}

Human motion generation (HMG) aims to translate a description into an action
whose poses are both semantically appropriate and coherent over time.
Most existing methods inject text semantics through cross-attention; others
pool language features into a global condition~\citep{tevet2023mdm,
zhang2024motiondiffuse,chen2023mld}. In conventional cross-attention,
motion queries read fixed text features, but the same interaction does not
update the text states. This one-way conditioning limits joint refinement
of text and motion tokens and motivates bidirectional modeling.

Multimodal diffusion transformers (MMDiTs) offer this interaction in visual
generation~\citep{esser2024sd3,kong2024hunyuanvideo}, making their transfer
to HMG a natural next step. Articulated motion, however, differs from visual
data in how its features relate: neighboring joints have distinct roles and
need not have similar orientations, while each joint must evolve coherently
over time. Captured motion accordingly has strong same-joint correlation
across adjacent frames ($0.96$) but weak same-frame correlation between
neighboring joints ($0.14$; \cref{tab:corr}).

A direct MMDiT transfer with flow matching does not preserve this motion
structure. It improves Speed but reduces Smoothness from $80.1$ to $52.5$
(\cref{tab:arch}), and its parent--child angular-speed correlation is
$0.25$, compared with $0.54$ in captured motion (\cref{tab:corr}). Thus
joint dynamics are poorly coordinated, producing jerky output despite
framewise-accurate individual poses.

We propose \textbf{\method{}}, a kinematics-aware MMDiT framework for HMG.
Fully shared attention updates text and motion tokens in both directions,
while flow matching predicts explicit body parameters. Geometric losses
supervise the articulated pose, and rotational-kinematics losses compare
actual root and joint rotations and their changes over time. A two-stage
curriculum progresses from broad motion learning to detailed caption
alignment with the same backbone and objective. Together, these components
combine bidirectional semantic interaction with supervision tailored to
coordinated, temporally coherent motion.

We construct a benchmark of $40{,}025$ held-out clips from six public
datasets, spanning diverse actions and six complementary evaluation
dimensions. With a common evaluator and scoring protocol, \method{}
surpasses Kimodo on five of six dimensions and improves the average
benchmark score by $40.8\%$ ($86.6$ vs.\ $61.5$). Kimodo retains the best
collision-based Plausibility score. Qualitative comparisons illustrate
action execution, while ablations examine the proposed supervision,
curriculum and data coverage within the framework.

We make three contributions:
\begin{itemize}
  \item \textbf{A kinematics-aware MMDiT.} We adapt multimodal diffusion
  transformers to HMG through fully shared bidirectional attention and
  flow matching over explicit body parameters.
  \item \textbf{Learning coherent, text-aligned motion.} Geometric and
  rotational-kinematics supervision addresses pose quality and coordinated
  temporal evolution; a two-stage curriculum refines broad motion learning
  with detailed captions.
  \item \textbf{A six-dataset benchmark.} We construct a common evaluation
  covering six axes, compare four released systems, and analyse the
  framework through correlation and training ablations.
\end{itemize}

%% file: sec/2_related.tex
\section{Related Work}
\label{sec:related}

\ifmoveoverflow

\paragraph{Text conditioning and motion representations.}
Text-to-motion methods include diffusion over continuous
features~\citep{tevet2023mdm,zhang2024motiondiffuse}, learned-latent
generation~\citep{petrovich2022temos,chen2023mld}, retrieval-augmented
diffusion~\citep{zhang2023remodiffuse}, and discrete-token
models~\citep{zhang2023t2mgpt,pinyoanuntapong2024mmm,guo2024momask,jiang2023motiongpt}.
Their pooled or cross-attention conditioning injects language into motion
features without jointly updating both token sequences.
Motion encoding also determines which structure can be supervised:
pelvis-relative features can obscure global quantities~\citep{meng2025acmdm},
whereas latent and discrete representations require a learned decoder before
body geometry is available. We retain explicit SMPL rotations, shape and root
motion so pose geometry and rotational changes remain accessible.
Geometric supervision is established in HMG~\citep{tevet2023mdm}; we
combine it with bidirectional text--motion interaction and explicit
supervision of how SMPL joint rotations change over time~\citep{loper2015smpl}.

\paragraph{Multimodal diffusion transformers.}
DiTs~\citep{peebles2023dit} and flow
matching~\citep{lipman2023flow,liu2023rectified} support joint multimodal
generation. Stable Diffusion~3~\citep{esser2024sd3} uses modality-specific
weights, while HunyuanVideo~\citep{kong2024hunyuanvideo} combines dual- and
single-stream blocks. HY-Motion~\citep{wen2025hymotion} also uses a hybrid
flow-matching DiT for motion. Our distinction is a fully shared stack
paired with geometric and rotational-kinematics supervision tailored to
articulated, temporally coherent motion. Joint attention can be bidirectional
while retaining modality-specific weights; sharing every block is a separate
architectural choice. Clean-data
prediction~\citep{li2026jit} exposes the body quantities used by these
losses. This complements larger-scale and
controllable generators~\citep{fan2025motionmillion,genmo2025,rempe2026kimodo,zhang2025motionanything}.

\paragraph{Training at scale and downstream use.}
Optical capture~\citep{mahmood2019amass} and language-labelled
corpora~\citep{plappert2016kitml,guo2022humanml3d,punnakkal2021babel}
provide motion supervision; video-derived
data~\citep{lin2023motionx,fan2025motionmillion,zhang2026romo} and staged
training~\citep{wen2025hymotion} broaden coverage.
ViMoGen~\citep{lin2025vimogen} transfers video-generation knowledge.
Our curriculum separates broad motion coverage from precise caption
supervision. Longer-sequence synthesis and humanoid
tracking~\citep{xiao2025motionstreamer,luo2026sonic,fu2026cosa05} are
downstream applications rather than our central contribution.

\else

\input{sec/mov/par_t2m}

\input{sec/mov/par_flowmatch}

\input{sec/mov/par_foundation}

\fi

%% file: sec/mov/par_t2m.tex
\paragraph{Text-to-motion.}
Text-to-motion methods use several representations and generative processes.
Continuous diffusion operates directly on motion
features~\citep{tevet2023mdm,zhang2024motiondiffuse}; latent
models~\citep{petrovich2022temos,chen2023mld} introduce a learned encoder
and decoder, and retrieval provides additional motion
references~\citep{zhang2023remodiffuse}. Discrete-token
models~\citep{zhang2023t2mgpt,pinyoanuntapong2024mmm,guo2024momask,jiang2023motiongpt}
quantise motion before generation. Recent work extends these directions to
causal streaming~\citep{xiao2025motionstreamer}, joint estimation and
generation~\citep{genmo2025}, multimodal
conditioning~\citep{zhang2025motionanything}, and controllable generation
at scale~\citep{rempe2026kimodo}. Our motivation concerns the distinction
between conditioning a motion backbone and jointly refining token-level
text and motion features, not the absence of language conditioning in
previous work.

The motion representation determines which constraints can be imposed
directly. ACMDM~\citep{meng2025acmdm} studies the limitations of
pelvis-relative HumanML3D features~\citep{guo2022humanml3d} and advocates
absolute coordinates. Our $151$-dimensional representation instead retains
SMPL rotations, shape and root motion. It exposes geometry and rotational
kinematics to supervision without a learned motion decoder. Earlier work
already supervises motion geometry~\citep{tevet2023mdm}; we combine this
principle with explicit rotational-kinematics targets in a shared MMDiT.
Body-frame root velocity removes dependence on the global starting
position, while direct rotations retain the body quantities needed by
the objective in \cref{sec:method:flow}.

%% file: sec/mov/par_flowmatch.tex
\paragraph{Multimodal transformers and flow matching.}
DiTs~\citep{peebles2023dit} replace convolutional denoisers with
transformers, while flow matching and rectified
flow~\citep{lipman2023flow,liu2023rectified} learn a continuous transport
from noise to data. Their combination has been successful in image and
video generation, with different choices for cross-modal interaction.
Stable Diffusion~3~\citep{esser2024sd3} uses joint attention with separate
modality parameters. HunyuanVideo~\citep{kong2024hunyuanvideo} first
processes modalities in dual-stream blocks and then combines them in shared
single-stream blocks. These examples motivate shared multimodal processing
without establishing that every layer must use one parameter set.

In motion, HY-Motion~\citep{wen2025hymotion} scales a hybrid flow-matching
DiT to a billion parameters. MotionMillion~\citep{fan2025motionmillion}
instead uses autoregressive discrete-token generation, while
Kimodo~\citep{rempe2026kimodo} and GENMO~\citep{genmo2025} use diffusion.
Efficient sampling has also been pursued through
distillation~\citep{dai2024motionlcm}. We therefore do not claim the first
MMDiT or flow-matching model for motion. \method{} instead studies how to
train a fully shared stack on continuous SMPL parameters for both semantic
alignment and temporal quality. Its clean-data prediction is related to
direct denoising parameterisations~\citep{li2026jit}, and exposes decoded
geometry and rotational kinematics to supervision. The cumulative ablation
in \cref{tab:arch} supports these motion-specific targets within this
framework; it neither isolates the effect of shared weights nor establishes
that flow matching is intrinsically less smooth than diffusion.

%% file: sec/mov/par_foundation.tex
\paragraph{Training at scale and downstream use.}
Motion corpora have expanded in both coverage and volume.
AMASS~\citep{mahmood2019amass} unifies optical capture under SMPL, and
language-labelled datasets provide text--motion
pairs~\citep{plappert2016kitml,guo2022humanml3d,punnakkal2021babel}.
Other collections cover dance, grasping, objects, scenes and individual
styles~\citep{li2021aist,taheri2020grab,li2023omomo,araujo2023circle,
jiang2024trumans,hassan2021samp,kim2025permo,luo2026sonic,mclean2025embody}.
Video reconstruction extends this coverage beyond the
studio~\citep{lin2023motionx,wang2024holistic,fan2025motionmillion,zhang2026romo},
although its geometric reliability differs from optical capture.
HY-Motion~\citep{wen2025hymotion} uses large-scale pretraining followed by
refinement, while ViMoGen~\citep{lin2025vimogen} transfers semantic knowledge
from a video generator. Staged learning is therefore not itself our novelty.
Our data design separates geometric validity from caption specificity:
long studio takes can teach motion kinematics without providing a precise
description of each short window. We add in-house capture and use separate
sampling policies for broad-coverage pretraining and sentence-level
finetuning (\cref{sec:data:tiers}).

Longer-sequence synthesis and robot execution are downstream applications,
not the distinguishing contributions of \method{}.
MotionStreamer~\citep{xiao2025motionstreamer} explores streaming through a
causal latent model, while whole-body control learns to track diverse motion
references~\citep{yuan2026bfm,luo2026sonic,fu2026cosa05}. Our generator
retains a fixed context and composes longer sequences through in-loop window
fusion. For execution, SMPL rotations and root motion serve as a kinematic
reference for robot-specific retargeting and tracking. These interfaces
reuse the learned representation without removing the need for local
sampling or embodiment-specific control. The examples establish feasibility
for the shown requests and two platforms, not broader performance claims.

%% file: sec/4_method.tex
\section{Method}
\label{sec:method}

\begin{figure}[t]
  \centering
  \includegraphics[width=\linewidth]{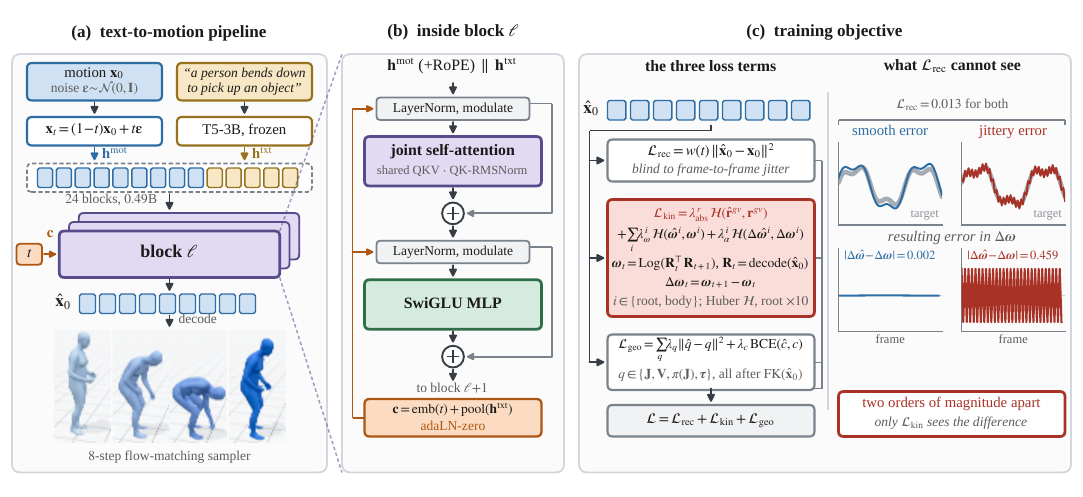}
  \caption{\textbf{Kinematics-aware multimodal generation.} \textbf{(a)} Text and motion
  tokens enter one $24$-block stack as a single sequence predicting
  clean body parameters $\hat{\mathbf{x}}_0$ directly.
  \textbf{(b)} A block shares
  QKV, MLP and adaLN~\citep{peebles2023dit} weights across modalities.
  \textbf{(c)} Reconstruction, rotational kinematics and decoded geometry
  provide complementary supervision. At right, predictions with equal
  reconstruction loss but different temporal errors, distinguished by
  $\mathcal{L}_{\text{kin}}$.}
  \label{fig:framework}
\end{figure}

\method{} adapts an MMDiT with flow matching to the structure of articulated
motion. Fully shared attention refines text and motion tokens
bidirectionally. Because weakly correlated joints must nevertheless evolve
coherently, geometric losses supervise the articulated pose and
rotational-kinematics losses supervise actual rotations and their changes
over time (\cref{fig:framework}). The two-stage curriculum in
\cref{sec:data} progresses from broad motion learning to detailed caption
alignment under the same backbone and objective.

\subsection{Bidirectional text--motion MMDiT}
\label{sec:method:arch}

\subsubsection{Motion representation}
\label{sec:method:repr}

\input{sec/mov/par_repr}

This compact representation is generated directly; forward kinematics
evaluates its body geometry without a learned motion decoder.
Root velocity removes dependence on the global starting position, and
gravity-aligned orientation supplies the root loss's vertical reference.
The continuous $6$D encoding represents each joint's rotation; it does not
imply similar coordinates for adjacent joints or smooth changes over time.
\cref{tab:corr} measures these distinct spatial and temporal relationships.
\input{sec/mov/par_stdise}
\input{sec/mov/par_contact}

\subsubsection{Bidirectional token interaction}

\input{sec/mov/par_dit}

\paragraph{Shared modulation.}
\input{sec/mov/par_adaln}

\subsection{Kinematics-aware flow matching}
\label{sec:method:flow}

\paragraph{Flow matching.}
We train with a rectified flow~\citep{liu2023rectified,lipman2023flow} between
the data distribution and $\mathcal{N}(0,\mathbf{I})$: for
$t \sim \mathcal{U}[0,1]$ and
$\boldsymbol{\epsilon} \sim \mathcal{N}(0,\mathbf{I})$ the interpolant is
linear,
\begin{equation}
  \mathbf{x}_t = (1-t)\,\mathbf{x}_0 + t\,\boldsymbol{\epsilon},
  \label{eq:interp}
\end{equation}
and the network predicts the clean sample
$\hat{\mathbf{x}}_0 = f_\phi(\mathbf{x}_t, t, \mathbf{h}^{\text{txt}})$
rather than the velocity field. For $t>0$, the corresponding velocity is
$\hat{\mathbf{v}}_t=(\mathbf{x}_t-\hat{\mathbf{x}}_0)/t$. A velocity model
could also recover a clean estimate; direct prediction makes it the network
output, which we convert to rotations and evaluate through forward kinematics.
The reconstruction term is
weighted by a truncated signal-to-noise ratio~\citep{hang2023minsnr},
\begin{equation}
  \mathcal{L}_{\text{rec}}
    = \mathbb{E}\Big[ w(t)\,\big\| \hat{\mathbf{x}}_0 - \mathbf{x}_0
      \big\|^2_{\mathcal{M}} \Big], \;\;
  w(t) = \frac{\min\!\big(\mathrm{SNR}(t), \gamma\big)}{Z},
  \label{eq:loss_rec}
\end{equation}
\input{sec/mov/par_minsnr}

\paragraph{Supervising temporal structure.}
Temporal attention exchanges information, but framewise reconstruction
does not directly constrain changes between poses. Slowly varying and
alternating errors in $6$D rotation space can receive equal penalties.
We therefore compare actual relative rotations on the $3$D rotation group
$\mathrm{SO}(3)$ and
their temporal changes, rather than only differences between coordinate
encodings. This addresses temporal behavior without treating adjacent
joints as interchangeable or forcing their rotations to agree.
Writing $\mathbf{R}_t$ for a joint's orientation at motion frame $t$,
define the body-frame angular increment and its difference,
\begin{equation}
  \boldsymbol{\omega}_t = \mathrm{Log}\!\big(\mathbf{R}_t^{\!\top}\mathbf{R}_{t+1}\big),
  \quad
  \Delta\boldsymbol{\omega}_t = \boldsymbol{\omega}_{t+1} - \boldsymbol{\omega}_t .
  \label{eq:so3vel}
\end{equation}
These targets describe rotational kinematics, not forces or torques.
For any fixed $\mathbf{G}\in\mathrm{SO}(3)$,
$(\mathbf{G}\mathbf{R}_t)^{\!\top}(\mathbf{G}\mathbf{R}_{t+1})
=\mathbf{R}_t^{\!\top}\mathbf{R}_{t+1}$, so the relative-rotation targets
are invariant to a constant reference rotation. The absolute root term
still supervises gravity-aligned orientation.

Writing $\mathcal{H}$ for the Huber loss~\citep{huber1964} with
$\beta = 0.05$,
applied to the gravity-aligned root ($r$) and each body joint ($j$),
\begin{equation}
\begin{aligned}
  \mathcal{L}_{\text{kin}}
  &= \lambda^{r}_{\text{abs}}\,\mathcal{H}\big(\hat{\mathbf{r}}^{gv}, \mathbf{r}^{gv}\big)
   + \lambda^{r}_{\text{vel}}\,\mathcal{H}\big(\hat{\boldsymbol{\omega}}^{r}, \boldsymbol{\omega}^{r}\big) \\
  &\quad + \lambda^{r}_{\text{acc}}\,\mathcal{H}\big(\Delta\hat{\boldsymbol{\omega}}^{r}, \Delta\boldsymbol{\omega}^{r}\big) \\
  &\quad + \frac{1}{J}\sum_{j=1}^{J} \Big[
      \lambda^{b}_{\text{vel}}\,\mathcal{H}\big(\hat{\boldsymbol{\omega}}^{j}, \boldsymbol{\omega}^{j}\big)
    + \lambda^{b}_{\text{acc}}\,\mathcal{H}\big(\Delta\hat{\boldsymbol{\omega}}^{j}, \Delta\boldsymbol{\omega}^{j}\big)
    \Big],
\end{aligned}
  \label{eq:loss_kin}
\end{equation}
\input{sec/mov/par_weights}
The total objective is
$\mathcal{L} = \mathcal{L}_{\text{rec}} + \mathcal{L}_{\text{kin}} +
\mathcal{L}_{\text{geo}}$, with per-term weights in \cref{sec:suppl:impl}.
Both training stages optimise it unchanged, with captions dropped
at probability $0.1$ to train the unconditional branch used for guidance.
The geometric terms constrain body poses, while rotational supervision
matches their changes to observed motion rather than suppressing movement.
Query--key normalisation, zero-initialised modulation and truncated loss
weighting stabilise optimisation. At inference, we use second-order
UniPC~\citep{zhao2023unipc} with $8$ steps and classifier-free
guidance~\citep{ho2022cfg} at scale $2.5$ for the quantitative flow-matching
results; exceptions are specified with their results.

%% file: sec/mov/par_repr.tex
We retain SMPL parameters to make body geometry available without a learned
motion decoder, and express the root by velocity rather than position to
avoid dependence on a window's arbitrary global starting point. A
motion is a sequence of $T$ frames at $30$\,fps, each a $151$-dimensional
vector
\begin{equation}
\begin{aligned}
  \mathbf{x}_t &= \big[\,
    \boldsymbol{\theta}_t \,\|\,
    \boldsymbol{\beta}_t \,\|\,
    \mathbf{r}^{c}_t \,\|\,
    \mathbf{r}^{gv}_t \,\|\,
    \mathbf{v}_t
  \,\big] \in \mathbb{R}^{151}, \\
  |\boldsymbol{\theta}_t|&=126,\;
  |\boldsymbol{\beta}_t|=10,\;
  |\mathbf{r}^{c}_t|=|\mathbf{r}^{gv}_t|=6,\;
  |\mathbf{v}_t|=3,
\end{aligned}
  \label{eq:repr}
\end{equation}
where $\boldsymbol{\theta}_t$ holds the $J=21$ body joints in the continuous
$6$D rotation parameterisation~\citep{zhou2019continuity},
$\boldsymbol{\beta}_t$ the shape coefficients, $\mathbf{r}^{c}_t$ and
$\mathbf{r}^{gv}_t$ the root orientation in the camera and in a
gravity-aligned frame, and $\mathbf{v}_t$ the body-frame root velocity,
integrated at decode time to recover translation.

%% file: sec/mov/par_stdise.tex
Each channel is standardised, with the standard deviation clamped at one so
near-constant channels are not amplified into noise.

%% file: sec/mov/par_contact.tex
Foot contact is supervised separately to keep the sampled state continuous:
six contact logits are predicted by an auxiliary head on the denoiser's
features instead of being sampled as binary channels in \cref{eq:repr}.
This shapes shared features without adding another sampled variable.

%% file: sec/mov/par_dit.tex
A frozen T5-3B~\citep{raffel2020t5} encoder represents each caption as at most
$50$ token embeddings, precomputed for training. Separate input projections
map text and motion into a common feature space, and the concatenated
sequence $[\mathbf{h}^{\text{mot}} ; \mathbf{h}^{\text{txt}}]$ passes through
$24$ transformer blocks of width $1024$ with $16$ attention heads, totalling
$495$M parameters. Every block jointly updates text and motion with shared
QKV projections, feed-forward weights and adaptive layer normalisation;
the output head reads only motion tokens. Joint attention lets text queries
read motion states as well as motion queries read text states. This differs
from dual-stream joint attention through parameter sharing, not bidirectionality
alone. Motion tokens use
rotary position embeddings~\citep{su2024roformer} to encode relative frame
positions, while text retains the contextual information supplied by its
encoder without assigning word indices to the motion time axis. Per-head
RMS normalisation of queries and keys~\citep{zhang2019rmsnorm,dehghani2023vit22b}
controls attention scale during bf16 training.

%% file: sec/mov/par_adaln.tex
Global conditioning supplements token-level joint attention rather than
replacing it. The flow time and mean-pooled caption enter through
adaLN-zero~\citep{peebles2023dit}: writing
$\mathbf{c} = \mathrm{emb}(t) + \mathrm{pool}(\mathbf{h}^{\text{txt}})$, each
block draws six modulation vectors from $\mathbf{c}$ and applies
\begin{align}
  \mathbf{h} &\leftarrow \mathbf{h} + \boldsymbol{\gamma}_1 \odot
     \mathrm{Attn}\!\big(\mathrm{mod}(\mathrm{LN}(\mathbf{h}),
     \boldsymbol{\mu}_1, \boldsymbol{\sigma}_1)\big), \nonumber \\
  \mathbf{h} &\leftarrow \mathbf{h} + \boldsymbol{\gamma}_2 \odot
    \mathrm{MLP}\!\big(\mathrm{mod}(\mathrm{LN}(\mathbf{h}),
     \boldsymbol{\mu}_2, \boldsymbol{\sigma}_2)\big),
  \label{eq:adaln}
\end{align}
with $\mathrm{mod}(\mathbf{h},\boldsymbol{\mu},\boldsymbol{\sigma}) =
(1+\boldsymbol{\sigma})\odot\mathbf{h} + \boldsymbol{\mu}$ and the gates
$\boldsymbol{\gamma}$ zero-initialised. The feed-forward is
SwiGLU~\citep{shazeer2020glu} of hidden width $2752$, and the output
projection is likewise zero-initialised. The zero gates suppress residual
updates, so the clean-motion output also begins at zero.

%% file: sec/mov/par_minsnr.tex
with $\mathrm{SNR}(t) = \big((1-t)/t\big)^2$, $\gamma = 5$, $Z$ a per-batch
normaliser and $\|\cdot\|_{\mathcal{M}}$ masked to valid frames. Without the
truncation the $t \to 0$ end of the schedule dominates the gradient and the
model spends capacity on a nearly trivial denoising problem.

%% file: sec/mov/par_weights.tex
with $\lambda^{r} = (5, 10, 10)$ and $\lambda^{b} = (1,1)$. Root kinematics
receive greater weight because root errors affect the whole body's motion.
The derivative terms target the observed kinematics rather than zero,
encouraging temporal coherence while matching the reference movement.
Complementary geometric terms supervise
body geometry through $3$D joint and vertex reconstruction, $2$D
reprojection, translation, and the contact cross-entropy described in
\cref{sec:method:repr}.

%% file: sec/3_data.tex
\section{Training Data and Curriculum}
\label{sec:data}

Geometric and rotational-kinematics supervision relies on reliable pose
trajectories, while bidirectional text--motion learning benefits from
precise captions. These properties do not always coincide. We filter motion
quality, then use a curriculum to accommodate differences in caption detail.

\subsection{Data sources}
\label{sec:data:corpus}

\input{sec/mov/fig_datamix}

\label{sec:data:ours}
Our data pool combines public optical motion capture and video-derived
motion~\citep{mahmood2019amass,fan2025motionmillion,zhang2026romo} with
our in-house data. It contains $1{,}983$ hours, counting mirrored
augmentation and held-out splits; \cref{tab:corpus,fig:datamix} detail the
sources and distribution. Our optical recordings contribute $1{,}187$ hours
under the same convention; neither total is unique training-only capture
time. Long in-house takes carry action labels, while short clips have
sentence-level descriptions. Take-level labels need not describe every
sampled window, motivating separate treatment from precise caption pairs.

\subsection{Quality control}
\label{sec:data:curation}

\ifmoveoverflow\else\input{sec/mov/par_canon}\fi

\ifmoveoverflow\else\input{sec/mov/par_takes}\fi

\ifmoveoverflow\else\input{sec/mov/par_gates}\fi

\ifmoveoverflow
All sources use a common body representation and coordinate convention.
Geometric filters reject severe skating, implausible poses and reconstruction
artefacts. Jitter filtering combines temporal coherence with speed-conditioned
thresholds to distinguish noise from fast valid motion. Separate rules remove
empty, invalid or out-of-scope captions without assuming equal caption
granularity (\cref{sec:suppl:jitter,sec:suppl:captions}).

\else
\input{sec/mov/par_jitter}
\fi

\ifmoveoverflow\else\input{sec/mov/par_align}\fi

\subsection{Coarse-to-fine curriculum}
\label{sec:data:tiers}

The two training tiers are disjoint. Stage~1 learns broad motion structure
from geometrically filtered data with heterogeneous captions, including
long in-house recordings, sampled by window count. Stage~2 starts from these
weights and refines alignment on six public sentence-labelled corpora and
our short recordings, with per-corpus and action-frequency caps. Both stages
retain text conditioning and the same geometric and rotational-kinematics
losses, and exclude evaluation clips. The schedule changes the supervision
distribution, not the task or representation. \cref{sec:exp:staging}
compares training recipes; \cref{sec:suppl:impl} gives their settings.

\ifmoveoverflow\else\input{sec/mov/par_recipe}\fi

%% file: sec/mov/fig_datamix.tex
\begin{figure}[t]
  \centering
  \includegraphics[width=\linewidth]{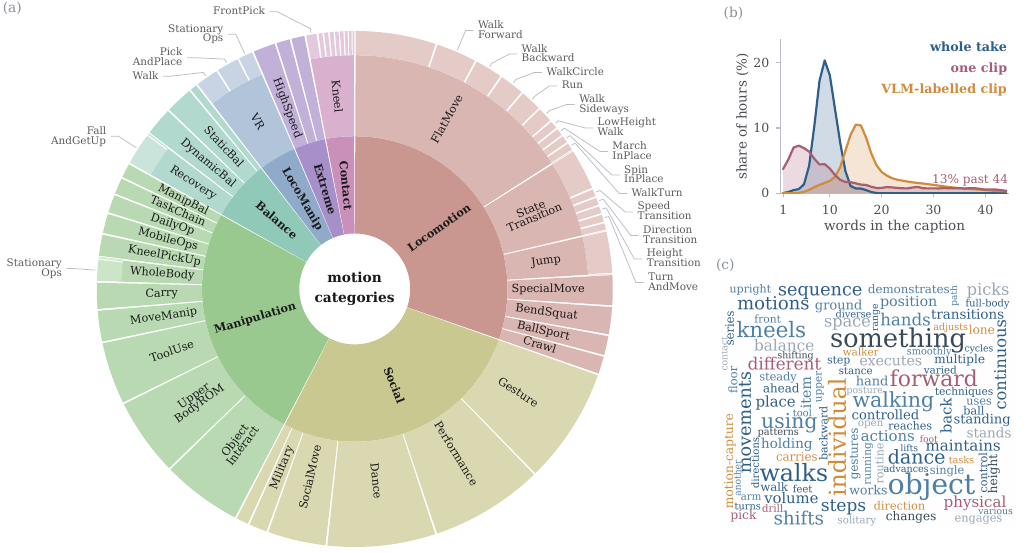}
  \caption{\textbf{Corpus composition.} \textbf{(a)} Motion categories over
  all $1{,}983$ hours; the three rings are family, subfamily and action, and
  arc is hours. \textbf{(b)} Caption length over all corpora, weighted by
  hours. \textbf{(c)} Our own captions' vocabulary. Hours include mirrored
  augmentation and held-out splits.}
  \label{fig:datamix}
\end{figure}

%% file: sec/mov/par_canon.tex
\paragraph{Canonicalisation and windowing.}
Clips are canonicalised with shape coefficients zeroed, the initial facing
aligned to $+X$, and the feet placed at $z=0$. In-house captures are retargeted
from the capture rig to SMPL~\citep{loper2015smpl} using local joint
rotations. Collar joints receive a bind-pose correction. This preserves
orientation information that a position-only target would not determine.

%% file: sec/mov/par_takes.tex
After conversion, our long in-house recordings (StudioTake) are stored as
complete performances and sampled through
random training windows. Their take-level labels provide action-category
conditioning rather than precise window-level descriptions. Sampling by the
number of available windows lets longer recordings contribute proportionally
without requiring a separate manual segmentation step.

%% file: sec/mov/par_gates.tex
\paragraph{Geometric filtering.}
A trusted subset of sentence-labelled motion provides reference
distributions for the geometric filters. We check jerk spikes, foot skating,
self-collision, body tilt and implausible jumps. Most thresholds are upper
or lower percentiles of the reference subset; clip length and skating use
absolute thresholds. This keeps the filtering criteria tied to reliable
motion while accounting for the different artefacts found across sources.

%% file: sec/mov/par_jitter.tex
\paragraph{Speed-aware jitter filtering.}
High-frequency energy can reflect either reconstruction noise or a fast
action. After low-pass filtering, we characterise the residual
$\mathbf{r}_t$ by lag-one correlation and acceleration sign changes,
\begin{equation}
\begin{aligned}
  \rho_1
    &= \frac{\sum_t \langle \mathbf{r}_t, \mathbf{r}_{t+1}\rangle}
            {\sum_t \|\mathbf{r}_t\|^2}, \\
  a_{\text{flip}}
    &= \frac{1}{T}\sum_t
       \mathbb{1}\big[\mathrm{sgn}(\ddot{\mathbf{r}}_t)
         \neq \mathrm{sgn}(\ddot{\mathbf{r}}_{t-1})\big].
\end{aligned}
  \label{eq:jitter}
\end{equation}
These scale-free statistics describe temporal coherence rather than
amplitude alone. Residual-magnitude thresholds are applied within speed
deciles, reducing the tendency to reject fast motion solely because its
residual is large.

%% file: sec/mov/par_align.tex
We do not apply a global threshold from a contrastive text--motion
model~\citep{petrovich2023tmr}: an alignment score can reflect domain
differences as well as annotation errors. Instead, explicit rules remove
invalid or refusal captions, descriptions shorter than two words, and
two-person or seated descriptions outside the selected training scope.
A caption-aware static check rejects clips whose text describes movement
while the body remains frozen. Surviving captions are cleaned, augmented
to three descriptions per clip where fewer exist, and encoded offline by
T5-3B~\citep{raffel2020t5}. These checks address unusable annotations without
treating a coarse but valid action label as a precise description of every
training window, preserving the distinction underlying the curriculum.

%% file: sec/5_experiments.tex
\section{Experiments}
\label{sec:exp}

We evaluate the two goals of the framework: text--motion alignment through
bidirectional modeling, and coordinated, temporally coherent motion under
the MMDiT--flow backbone. A common six-dataset benchmark compares complete
systems; correlation analysis and ablations examine the motion-specific
supervision, curriculum and data scale. \cref{sec:exp:deployment} shows
downstream use.

\subsection{Experimental setting}
\label{sec:exp:protocol}

\ifmoveoverflow
We construct our benchmark from $40{,}025$ held-out clips in six public
datasets: HumanML3D, BABEL, AIST++, GRAB, PerMo and BONES
(\cref{tab:corpus}), covering diverse actions and capture sources.
These clips are excluded from both training stages.
The combined split, evaluator and score mappings are constructed for this
study, rather than taken from an established community benchmark.
The evaluation sources are public; our full training recipe also uses
in-house capture.
We report six axes on fixed $0$--$100$ scales, higher being better.
\textbf{Text-Motion Match}, \textbf{Naturalness} and \textbf{Diversity} live
in the embedding space of a contrastive text--motion
evaluator~\citep{petrovich2023tmr}, following text--motion retrieval and
distributional evaluation~\citep{guo2022humanml3d,heusel2017fid}.
\textbf{Smoothness} combines whole-body, root and transition statistics
of third-order temporal differences of the angular increment in \cref{eq:so3vel};
\textbf{Plausibility} measures limb-capsule interpenetration. Both are
calibrated using captured motion rather than treating zero as the only
reference. Plausibility is a geometric
proxy, not a test of balance, dynamics or robot executability.
\textbf{Speed} combines first-output latency and amortised per-segment time
at batch size one. All methods use the same evaluator and scoring rules;
raw results and mappings are in \cref{sec:suppl:protocol}. We report one
generation seed and no significance analysis.
\else
\label{sec:suppl:protocol}
\input{sec/mov/par_axes}

\input{sec/mov/par_smooth}

\input{sec/mov/par_protocol}
\fi

\subsection{Generation quality}
\label{sec:exp:main}

\begin{table}[!htbp]
\centering\scriptsize
  \caption{\textbf{Comparison on our six-dataset benchmark.}
  Released multi-corpus systems are re-evaluated, not quoted from their
  papers. Training data and representations differ; ours includes in-house
  capture. Axes are custom $0$--$100$ scores, higher better; differences
  and relative gains refer to these scores, not the underlying raw metrics.
  \emph{Params} excludes frozen text encoders. \emph{Avg.}\ is the
  unweighted mean of the six scores.}
  \label{tab:main}
  \setlength{\tabcolsep}{3pt}
  \begin{tabular}{llrccccccc}
    \toprule
    Method & Venue & Params & Nat.\,$\uparrow$ & Match\,$\uparrow$
           & Div.\,$\uparrow$ & Smth.\,$\uparrow$ & Speed\,$\uparrow$
           & Plaus.\,$\uparrow$ & Avg.\,$\uparrow$ \\
    \midrule
    MotionMillion~\citep{fan2025motionmillion} & ICCV'25 & $7.8$B & 42.3 & 30.5 & 51.7 & 49.5 & 45.5 & 72.1 & 48.6 \\
    HY-Motion~\citep{wen2025hymotion} & arXiv'25 & $1.0$B & 38.0 & 52.6 & 67.3 & 54.3 & 31.4 & 71.4 & 52.5 \\
    GENMO~\citep{genmo2025}          & ICCV'25 & $523$M & 44.2 & 28.0 & 52.3 & 79.8 & 20.6 & 68.1 & 48.8 \\
    Kimodo~\citep{rempe2026kimodo}   & arXiv'26 & $283$M & 36.8 & 51.0 & 59.1 & 79.0 & 49.5 & \textbf{93.4} & 61.5 \\
    \midrule
    \method (ours)                  &         & $495$M & \textbf{86.4} & \textbf{86.8} & \textbf{89.3}
                                    & \textbf{90.7} & \textbf{81.4} & 85.0 & \textbf{86.6} \\
    \bottomrule
  \end{tabular}
\end{table}

\input{sec/mov/par_scope}
\method{} surpasses Kimodo on five of six dimensions and raises the
average score from $61.5$ to $86.6$, a $40.8\%$ relative improvement
(\cref{fig:radar,tab:main}). The strongest baseline differs by dimension:
Text-Motion Match exceeds HY-Motion by $34.2$ points ($86.8$ vs.\ $52.6$),
and Smoothness exceeds GENMO by $10.9$ points ($90.7$ vs.\ $79.8$).
All \method{} scores use the same eight-step configuration. Plausibility
remains $8.4$ points below Kimodo ($85.0$ vs.\ $93.4$), so improvements
in alignment and temporal quality do not eliminate geometric failures.
The relative gain concerns the average of our mapped scores; raw results
and mappings are provided in \cref{sec:suppl:protocol}.

\paragraph{Qualitative comparison.}
\label{sec:exp:qual}

\input{sec/mov/fig_qualitative}

\cref{fig:qualitative} makes the semantic differences more concrete by
comparing the same prompts across the methods in \cref{tab:main}.
\ifmoveoverflow\else\input{sec/mov/par_prompts}\input{sec/mov/par_grid}\fi
\method{} descends and returns to standing for the pickup prompt, and
maintains a leftward arm extension for the gesture prompt. These selected
trajectories illustrate action execution; aggregate retrieval scores assess
alignment across the evaluation pool. \cref{sec:suppl:qual} gives more examples.

\subsection{Representation analysis and ablations}
\label{sec:exp:ablation}

We examine rotational supervision and inter-joint coordination before
the curriculum and data coverage.
\cref{tab:arch,tab:scaling} use single-stage training on the six public
fine-tier corpora only, rather than the full recipe of \cref{tab:main}.

\subsubsection{Backbone and temporal supervision}
\label{sec:exp:arch}

\begin{table}[t]
  \centering
  \captionsetup{font=footnotesize,justification=raggedright,singlelinecheck=false,skip=4pt}
  \renewcommand{\arraystretch}{1.08}
  \begin{minipage}[t]{0.52\linewidth}
    \caption{\textbf{Architecture and objective.} Cumulative single-stage
    ablations. DDIM~\citep{song2021ddim}: RoPE encoder; MMDiT+flow:
    shared multimodal DiT with rectified flow. Avg.\ excludes Speed; higher is better.}
    \label{tab:arch}
    \centering\scriptsize
    \setlength{\tabcolsep}{1.5pt}
    \begin{tabular*}{\linewidth}{@{\extracolsep{\fill}}lccccccc@{}}
      \toprule
      Config. & Nat. & Match & Div. & Smth. & Speed & Plaus. & Avg. \\
      \midrule
      DDIM, RoPE       & 75.6 & 80.8 & 85.3 & 80.1 & 41.1 & 70.8 & 78.5 \\
      MMDiT+flow       & 81.4 & 80.8 & 86.3 & 52.5 & 81.5 & 81.9 & 76.5 \\
      \;\; + mirroring  & \textbf{83.2} & 83.2 & 88.8 & 59.5 & 81.0 & 82.9 & 79.5 \\
      \;\; + root rot.  & 81.1 & 82.7 & \textbf{89.5} & 74.8 & \textbf{81.6} & \textbf{89.3} & 83.5 \\
      \;\; + body rot.  & 82.6 & \textbf{87.3} & 86.1 & \textbf{90.9} & 78.3 & 87.0 & \textbf{86.8} \\
      \bottomrule
    \end{tabular*}
  \end{minipage}\hfill
  \begin{minipage}[t]{0.45\linewidth}
    \caption{\textbf{Motion correlations.} Top: temporal coherence and weak
    same-frame joint correlation in captured motion. Bottom: parent--child
    angular-speed coordination.}
    \label{tab:corr}
    \centering\scriptsize
    \setlength{\tabcolsep}{2pt}
    \begin{tabular*}{\linewidth}{@{\extracolsep{\fill}}lc@{}}
      \toprule
      \emph{Data neighbors} & Joint correlation \\
      Same joint, adjacent frames & 0.96 \\
      Adjacent joints, same frame & 0.14 \\
      \midrule
      \emph{Coordination} & Angular speed correlation \\
      Captured motion & 0.54 \\
      MMDiT+flow & 0.25 \\
      \method{} (ours) & \textbf{0.44} \\
      \bottomrule
    \end{tabular*}
  \end{minipage}
\end{table}

The direct-transfer configuration in \cref{tab:arch} replaces the RoPE
encoder and DDIM with the shared MMDiT and rectified flow. It raises Speed
from $41.1$ to $81.5$, but reduces Smoothness from $80.1$ to $52.5$.
Mirroring improves Smoothness to $59.5$, still below the diffusion baseline.
Adding root rotational supervision raises it by $15.3$ points to $74.8$;
adding body rotational supervision contributes a further $16.1$ points,
reaching $90.9$. Across the two supervision additions, Text-Motion Match
increases from $83.2$ to $87.3$. Matching rotational kinematics therefore
improves temporal quality without sacrificing the requested action in this
comparison. Not every score increases: Plausibility falls from $89.3$ to
$87.0$ after body supervision. The first replacement changes both
architecture and generative process; the cumulative rows support the full
kinematics-aware objective, not the isolated effects of bidirectional
attention, flow matching or individual losses.

\paragraph{Temporal and inter-joint correlations.}
\Cref{tab:corr} computes Pearson $r$~\citep{pearson1895regression} per clip
on axis--angle rotation magnitudes: same-joint adjacent frames and non-root
parent--child pairs within a frame; coordination uses parent--child
body-frame angular-speed magnitudes. We average over joints or pairs and
clips ($1{,}481$ captures; $1{,}500$ generations per model, seed~$0$);
\method{} is the full two-stage model in \cref{tab:main}. Captured motion is
more temporally than cross-joint correlated ($0.96$ vs.\ $0.14$).
\method{} raises generated angular-speed correlation from $0.25$ to
$0.44$, towards capture's $0.54$. Together with the Smoothness results in
\cref{tab:arch}, this characterises the direct transfer as poorly coordinated
and jerky; the correlations are descriptive rather than causal.

\input{sec/mov/tab_ablation_pair}
\input{sec/mov/sub_staging}
\input{sec/mov/sub_scaling}

\subsection{Downstream applications}
\label{sec:exp:deployment}

\paragraph{Long-sequence generation.}
We compose twelve prompts into a $36$-second motion containing locomotion,
gestures and posture changes (\cref{fig:longseq}). Overlapping eight-second
windows share predictions within each sampling step, reusing the trained
generator without increasing its attention context. This application
illustrates action composition, not a quantitative assessment of drift;
\cref{sec:suppl:fuse,sec:suppl:longseq} give details and additional examples.

\input{sec/mov/fig_longseq}

\begin{figure}[!htbp]
  \centering
  \includegraphics[width=\linewidth]{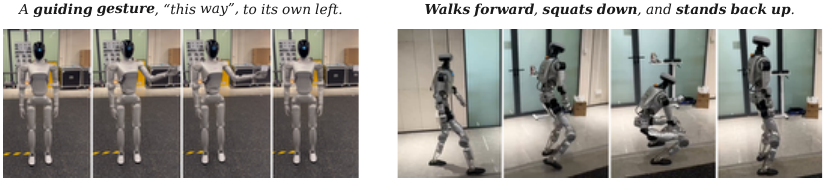}
  \caption{\textbf{Execution on two physical humanoids.} Left: LimX Luna
  performs a guiding gesture. Right: LimX Oli walks, squats and stands.
  Generated references are retargeted and tracked by the robot controller;
  \cref{fig:suppl:robot} provides the full examples.}
  \label{fig:robot}
\end{figure}

\paragraph{Humanoid execution.}
We retarget four generated motions to LimX Luna~\citep{limx_luna} and
LimX Oli~\citep{limx_oli}, then track them with the controller of
\citet{fu2026cosa05} (\cref{fig:robot,fig:suppl:robot}). The generator
provides a kinematic reference; robot-specific retargeting and control
handle execution. These demonstrations show feasibility on two platforms,
not a deployment success rate or a guarantee of dynamic stability.

\FloatBarrier

%% file: sec/mov/par_axes.tex
Our protocol reports six axes, mapped to $0$--$100$ by common scoring
rules. These protocol-specific scores do not have a universal scale.
Three use the embedding space of
a contrastive text--motion evaluator~\citep{petrovich2023tmr}:
\textbf{Text-Motion Match} (R@1/2/3 within batches of $32$, plus multimodal
distance, following~\citet{guo2022humanml3d}), \textbf{Naturalness}
(Fr\'echet distance~\citep{heusel2017fid} between generated and
ground-truth embedding statistics), and \textbf{Diversity} (mean pairwise
distance, scored by closeness to the ground-truth value rather than by
magnitude). \textbf{Plausibility} measures limb-capsule interpenetration
after forward kinematics; it is a proxy for geometric plausibility, not
a measure of balance, dynamic feasibility or hardware safety.
\textbf{Speed} combines first-output latency and amortised per-segment time
at batch size one. Per-axis score differences are reported in points.
For the average score $\bar{S}=\frac{1}{6}\sum_{k=1}^{6}S_k$, the relative
gain is $100(\bar{S}_{\mathrm{ours}}/\bar{S}_{\mathrm{base}}-1)\%$.
The reported $40.8\%$ improvement over Kimodo uses this composite score;
it is not a percentage change in raw retrieval, distance, collision or
latency quantities, and depends on the mappings below.

\paragraph{Fixed score mappings.}
Every raw metric is mapped to $0$--$100$ with constants fixed across
methods. Naturalness rewards lower segment and transition FID.
Text-Motion Match combines retrieval accuracy and text--motion distance,
while Diversity rewards proximity to the held-out capture distribution.
Plausibility decreases monotonically with excess interpenetration relative
to captured motion, whose legitimate body contact can activate this proxy.
Speed combines first-output and amortised per-segment latency against fixed
budgets at batch size one. Component scores are averaged within each axis.
These mappings make heterogeneous measurements comparable for this
protocol, but do not turn them into universal metrics or isolate differences
in model interfaces.

%% file: sec/mov/par_smooth.tex
\paragraph{Smoothness.}
This axis averages four statistics of third-order finite differences of
the body-frame angular increment $\boldsymbol{\omega}_t$ in \cref{eq:so3vel}:
mean whole-body magnitude, mean root magnitude, peak magnitude over
transition windows (PJ), and accumulated deviation from the reference
transition level (AUJ). These are frame-based statistics, not derivatives
in physical time. We derive fixed targets by low-pass filtering the held-out
capture pool to reduce measurement noise while retaining human movement.
Each component rewards closeness to its target, so both excessive jitter
and over-damped or near-frozen motion are penalised. Smoothness is the mean
of the four component scores. Its absolute values depend on this
protocol-specific calibration; lower uncalibrated temporal differences do
not necessarily indicate better motion.

%% file: sec/mov/par_protocol.tex
Our six-dataset benchmark comprises $40{,}025$ clips from public
corpora: HumanML3D, BABEL, AIST++, GRAB, PerMo and BONES. The pool is held
out from both of our training stages. Its public sources do not make the
combined split or our scoring rules an established benchmark. All methods
use the same six-corpus text--motion evaluator; this controls the evaluation
procedure, not differences in their training data. Our held-out designation
does not establish that these clips were unseen by every released baseline.
The \method{} row in \cref{tab:main} uses the full
two-stage recipe: a $495$M generator sampled with $8$ UniPC steps and
guidance $2.5$ (\cref{sec:suppl:impl}). The ablations use the training
scope and model settings specified in each table; in particular,
\cref{tab:arch,tab:scaling} use single-stage public-data training.
Qualitative long-sequence examples are separate from this evaluation, with
their configuration differences documented in \cref{sec:suppl:longseq}.

The reported runs use one generation seed (seed $0$), $30$\,fps motion and
retrieval groups of $32$. Text embeddings are aggregated by the evaluator's
mean-caption rule. We do not report confidence intervals or claim
statistical significance from these single-seed results. The score
calibration uses statistics of the held-out population; it is not an
independent validation of the scoring functions. To separate measured
differences from the choice of normalisation, \cref{tab:raw} reports the
underlying retrieval, distance and collision values.

\begin{table}[htbp]
	\centering\scriptsize
	\caption{\textbf{Raw metrics underlying the main comparison.} R@1/2/3
	are percentages within $32$-candidate retrieval groups. Fr\'echet
	distances (FID) and text--motion distance (MM-Dist) use the
	same six-corpus evaluator for
	every row, so they are not directly comparable to numbers from a
	differently trained evaluator. Pen.\ is the capsule-interpenetration
	proxy, multiplied by $10^3$ for readability, not a collision frequency.}
	\label{tab:raw}
	\setlength{\tabcolsep}{3.5pt}
	\begin{tabular}{lrrrrrrr}
		\toprule
		Method & R@1$\uparrow$ & R@2$\uparrow$ & R@3$\uparrow$
		& Seg.\ FID$\downarrow$ & Trans.\ FID$\downarrow$
		& MM-Dist$\downarrow$ & Pen.$\downarrow$ \\
		\midrule
		MotionMillion & 25.3 & 36.9 & 44.3 & 0.201 & 0.409 & 1.237 & 13.31 \\
		HY-Motion & 38.9 & 54.0 & 62.7 & 0.233 & 0.478 & 1.187 & 13.90 \\
		GENMO & 22.7 & 34.6 & 43.0 & 0.212 & 0.338 & 1.236 & 16.83 \\
		Kimodo & 37.4 & 53.1 & 62.6 & 0.319 & 0.364 & 1.199 & \textbf{1.42} \\
		\method{} & \textbf{62.1} & \textbf{77.6} & \textbf{85.2}
		& \textbf{0.053} & \textbf{0.034} & \textbf{1.055} & 4.93 \\
		\bottomrule
	\end{tabular}
\end{table}

%% file: sec/mov/par_scope.tex
\cref{tab:main} compares multi-corpus models using their released checkpoints
and our common evaluator~\citep{petrovich2023tmr}. It compares available
systems rather than retraining them with matched data or compute. In
particular, our complete recipe includes non-public capture; the common
test set alone does not control for that difference.

%% file: sec/mov/fig_qualitative.tex
\begin{figure}[t]
  \centering
  \includegraphics[width=\linewidth]{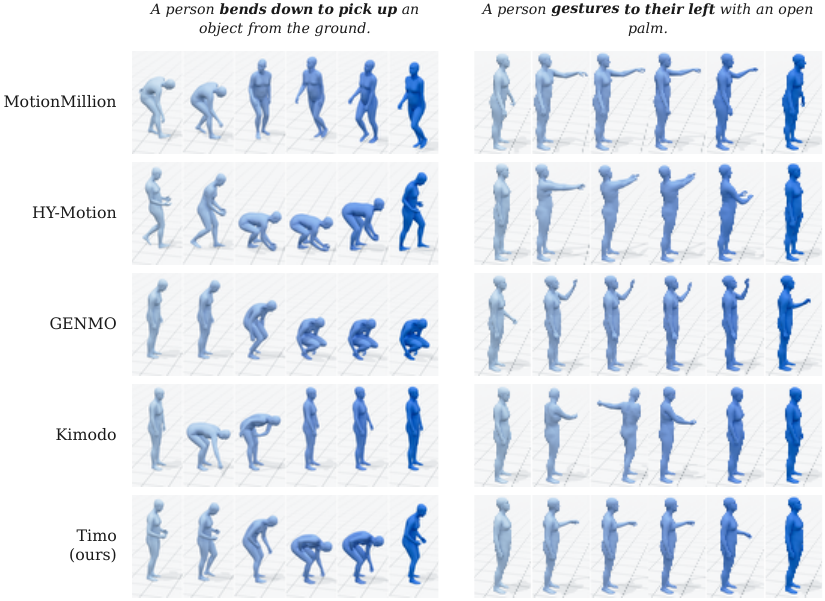}
  \caption{\textbf{Qualitative comparison.} One model per row, one prompt per
  column, moments left to right, shaded light to dark as time advances; a
  column shares one camera, so a body that sits lower or reaches further has
  really done so. \cref{fig:suppl:qual-step} carries two further prompts.}
  \label{fig:qualitative}
\end{figure}

%% file: sec/mov/tab_ablation_pair.tex
\begin{table}[t]
  \centering
  \captionsetup{font=footnotesize,justification=raggedright,singlelinecheck=false,skip=4pt}
  \renewcommand{\arraystretch}{1.08}
  \begin{minipage}[t]{0.52\linewidth}
    \caption{\textbf{Training curriculum.} Full-recipe comparison:
    public-only single-stage vs.\ two-stage training with in-house capture.
    Avg.\ excludes Speed; higher is better.}
    \label{tab:staging}
    \centering\scriptsize
    \setlength{\tabcolsep}{1.5pt}
    \begin{tabular*}{\linewidth}{@{\extracolsep{\fill}}lcccccc@{}}
      \toprule
      Recipe & Nat. & Match & Div. & Smth. & Plaus. & Avg. \\
      \midrule
      1 stage, fine      & 82.6 & \textbf{87.3} & 86.1 & \textbf{90.9} & \textbf{87.0} & 86.8 \\
      \;\; + coarse tier & 77.2 & 73.8 & 86.9 & 90.2 & 86.2 & 82.8 \\
      2 stages (ours)    & \textbf{86.4} & 86.8 & \textbf{89.3} & 90.7 & 85.0 & \textbf{87.6} \\
      \bottomrule
    \end{tabular*}
  \end{minipage}\hfill
  \begin{minipage}[t]{0.45\linewidth}
    \caption{\textbf{Data scaling.} Fixed architecture and schedule;
    each of the six public corpora is uniformly subsampled.
    Avg.\ excludes Speed; higher is better.}
    \label{tab:scaling}
    \centering\scriptsize
    \setlength{\tabcolsep}{1.5pt}
    \begin{tabular*}{\linewidth}{@{\extracolsep{\fill}}lcccccc@{}}
      \toprule
      Train & Nat. & Match & Div. & Smth. & Plaus. & Avg. \\
      \midrule
      $1/8$  & 57.8 & 58.0 & 70.9 & 68.7 & 76.8 & 66.5 \\
      $1/4$  & 68.5 & 70.3 & 75.5 & 84.5 & 87.1 & 77.2 \\
      $1/2$  & 76.2 & 77.1 & 81.3 & 87.5 & \textbf{88.5} & 82.1 \\
      full   & \textbf{82.6} & \textbf{87.3} & \textbf{86.1} & \textbf{90.9} & 87.0 & \textbf{86.8} \\
      \bottomrule
    \end{tabular*}
  \end{minipage}
\end{table}

%% file: sec/mov/sub_staging.tex
\subsubsection{Data curriculum and coverage}
\label{sec:exp:staging}

Adding the coarse tier to single-stage training reduces Naturalness by
$5.4$ points ($82.6$ to $77.2$) and Text-Motion Match by $13.5$ points
($87.3$ to $73.8$; \cref{tab:staging}). Under this capped sampling policy,
adding data alone did not improve either metric. The complete two-stage recipe
recovers Naturalness to $86.4$ and Match to $86.8$. Relative to fine-only
training, it improves Naturalness by $3.8$ points but not Match, Smoothness
or Plausibility. This supports the full recipe as a way to use heterogeneous
data, not a uniform gain on every metric. Because the final row also adds
in-house capture and training steps, the comparison does not isolate the
effect of staging or control for compute.

%% file: sec/mov/sub_scaling.tex
\label{sec:exp:scaling}

At fixed architecture and training schedule, increasing each public
fine-tier corpus from $1/8$ to its full size raises Text-Motion Match from
$58.0$ to $87.3$, Naturalness from $57.8$ to $82.6$, and Smoothness from
$68.7$ to $90.9$ (\cref{tab:scaling}). Plausibility instead peaks at the
half-data setting ($88.5$ versus $87.0$ with full data). Together with the
curriculum comparison, this distinguishes the benefit of more examples
within one caption tier from mixing tiers with different supervision
quality. Neither experiment replaces the test of rotational supervision
in \cref{tab:arch}, which holds the corpus mixture fixed.

%% file: sec/mov/fig_longseq.tex
\begin{figure}[!htbp]
  \centering
  \includegraphics[width=\linewidth]{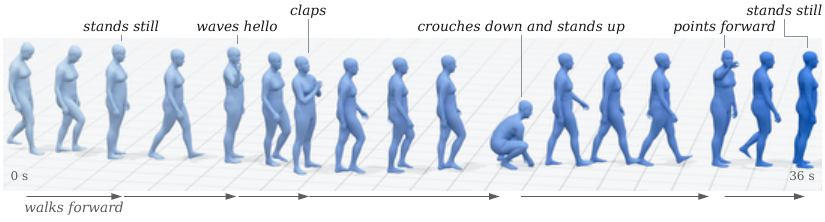}
  \caption{\textbf{Long-sequence application: twelve prompts over $36$ seconds.} Travelling and
  in-place prompts alternate, and \method{} generates them as a single motion,
  drawn in one scene with a common camera and scale. Displayed poses are
  aligned to the floor and shaded light to dark with time. Arrows span the
  travelling prompts.}
  \label{fig:longseq}
\end{figure}

%% file: sec/6_conclusion.tex
\section{Conclusion}
\label{sec:conclusion}

\method{} adapts bidirectional MMDiT processing and flow matching to
articulated motion, whose joints are weakly correlated in pose but must
evolve coherently. Whereas direct transfer yields poorly coordinated,
jerky dynamics, shared attention enables token-level text--motion
interaction, geometric losses constrain the decoded body, and
rotational-kinematics losses match how each joint turns and changes over
time. Coarse-to-fine training separates broad motion learning from detailed
caption alignment without changing this objective. Together, these
components retain efficient flow sampling while improving pose quality,
Smoothness and parent--child coordination. Longer-sequence generation and
humanoid execution reuse the same representation without modifying the
generator.

%% file: sec/7_statements.tex

\subsection*{AI use statement}

Generative AI tools assisted with manuscript polishing and literature lookup.
The
authors are responsible for verifying AI-assisted content, including
technical statements, citations and interpretations, and for the final text,
claims and artifacts.

\subsection*{Ethics statement}

Our in-house optical capture was recorded under written agreements covering
recording and research use. The retained data contains marker trajectories,
not video, audio or names, but gait and distinctive gestures may still
carry identifying information. Public corpora are cited individually and
used under their respective licences. Robot demonstrations were conducted
in a controlled laboratory with an operator able to stop execution.
Neither the generated references nor the collision-based Plausibility
metric certify safe execution on a robot. Generated motion could also
support misleading synthetic footage. Using a generic skeleton without
appearance or identity does not eliminate that risk.

\subsection*{Reproducibility statement}

\cref{sec:method} specifies the representation, shared backbone and
motion-aware objective; \cref{sec:suppl:impl} gives optimisation,
curriculum and sampling settings. Data sources and curation are described
in \cref{sec:data,sec:suppl:corpora}. Our evaluation is author-constructed:
\cref{sec:suppl:protocol} documents the pool, raw results and custom score
mappings. Released baseline checkpoints are re-evaluated with the same
evaluator, not retrained with matched data or compute. Reported runs use
one generation seed, without confidence intervals. We will release the
trained checkpoint, the six-corpus evaluator, and sampling and scoring
scripts. Public corpora remain available from their original authors;
our in-house capture cannot be redistributed, limiting exact reproduction
of the full training recipe. Public-data ablations address a narrower
training setting than that of the complete system.

%% file: sec/X_suppl.tex
\clearpage
\appendix
\raggedbottom
\setcounter{table}{0}
\setcounter{figure}{0}
\renewcommand{\thetable}{S\arabic{table}}
\renewcommand{\thefigure}{S\arabic{figure}}
\renewcommand{\theHtable}{S\arabic{table}}
\renewcommand{\theHfigure}{S\arabic{figure}}

\noindent
\textbf{Appendix A} gives architecture, optimisation and downstream
implementation details. \textbf{Appendix B} describes the training corpora
and data curation. \textbf{Appendix C} defines the evaluation protocol and
score mappings. \textbf{Appendix D} presents long-sequence, qualitative,
coverage and physical-deployment results. \textbf{Appendix E} provides
extended related work.

\section{Implementation details}
\label{sec:suppl:impl}

\subsection{Architecture and optimisation}

\cref{tab:suppl:settings} collects the settings used for the quantitative
experiments. The network shares QKV, feed-forward and adaLN parameters
across modalities as described in \cref{sec:method:arch}. Its motion head
predicts the explicit $151$-dimensional state, without a learned motion
encoder or decoder. Geometric and rotational-kinematics losses are evaluated
on the clean prediction in both training stages. Text embeddings
are precomputed with frozen T5-3B~\citep{raffel2020t5}; the encoder is not
finetuned or included in the parameter count. Training uses
PyTorch~\citep{paszke2019pytorch} distributed data parallelism and gradient
checkpointing~\citep{chen2016checkpoint}.

\begin{table}[htbp]
  \centering\small
  \caption{\textbf{Model and optimisation settings} shared by both training stages.}
  \label{tab:suppl:settings}
  \begin{tabular}{@{}lp{0.62\linewidth}@{}}
    \toprule
    Setting & Value \\
    \midrule
    Backbone & $24$ blocks, width $1024$, $16$ heads, $495$M parameters \\
    Feed-forward & SwiGLU, hidden width $2752$ \\
    Motion & $151$ features at $30$\,fps; $240$ frames per window \\
    Text & At most $50$ token embeddings of width $1024$ \\
    Optimiser & AdamW~\citep{loshchilov2019adamw}, learning rate $2\times10^{-4}$ \\
    Schedule & $100$k steps per stage; rate halved at $50$k and $80$k \\
    Batch & $64$ windows per GPU; $1024$ globally on $16$ H100 GPUs \\
    Numerical settings & Mixed bf16; gradient clipping $0.5$; no EMA \\
    Caption dropout & $0.1$ \\
    Sampler & Second-order UniPC, $8$ steps, uniform time grid \\
    Guidance & Classifier-free guidance scale $2.5$ \\
    Window fusion & $240$-frame windows, $60$-frame overlap \\
    \bottomrule
  \end{tabular}
\end{table}

\subsection{Curriculum and loss settings}

For the in-house data, we use the names \textbf{StudioTake} for long,
take-labelled recordings and \textbf{StudioClip} for short,
sentence-labelled recordings (\cref{tab:corpus}). Stage~1 samples the coarse
tier in proportion to its window counts, without
the fine tier's per-corpus or action-frequency caps. The older StudioTake
pass retains its $15$k-entry limit. Stage~2 loads only the learned weights,
restarting the optimiser and schedule. It uses the fine tier with a $30$k
per-corpus cap, a $15\%$ in-batch locomotion cap and $5\times$ upsampling of
StudioClip. There is no coarse-tier replay.

\begin{table}[htbp]
  \centering\small
  \caption{\textbf{Objective weights.} The objective is unchanged between
  stages. Huber terms use $\beta=0.05$; contact labels use a foot-velocity
  threshold of $0.15$\,m/s.}
  \label{tab:suppl:losses}
  \begin{tabular}{@{}llr@{}}
    \toprule
    Group & Term & Weight \\
    \midrule
    Reconstruction & Clean-motion error; min-SNR truncation $\gamma=5$ & $1$ \\
    Root rotation & Absolute / velocity / acceleration & $5\,/\,10\,/\,10$ \\
    Body rotation & Velocity / acceleration & $1\,/\,1$ \\
    Geometry & 3D joints / vertices & $500\,/\,500$ \\
     & 2D reprojection & $1000$ \\
     & Translation / contact BCE & $1\,/\,1$ \\
    \bottomrule
  \end{tabular}
\end{table}

\subsection{Downstream sampling and retargeting}
\label{sec:suppl:fuse}
\label{sec:method:window}

Training uses $240$-frame ($8$\,s) windows. For longer requests, we use
windows overlapping by $60$ frames, keeping the backbone context fixed.
Predictions are fused inside each sampling step, with per-segment captions
re-indexed onto each crop. This downstream procedure changes the number
of window evaluations, not the learned architecture; it is not a separate
contribution in flexible-duration generation.
\input{sec/mov/par_fuse}
\input{sec/mov/par_bitident}

\label{sec:method:robot}
For robot deployment, SMPL output is a kinematic reference rather than a
sequence of platform-specific commands. We retarget it to each robot and
track it with the whole-body controller of \citet{fu2026cosa05}.
Retargeting and control, not the unchanged generator, account for robot
morphology, joint limits and execution dynamics.

\FloatBarrier
\section{Additional data details}
\label{sec:suppl:corpora}

\subsection{Corpora}

\cref{tab:corpus} expands the source breakdown in \cref{fig:datamix},
distinguishing optical capture from video reconstruction and sentence-level
descriptions from labels covering a whole take. The two in-house subsets
serve the complementary roles described in \cref{sec:data}: StudioTake
contributes motion coverage to the coarse tier, while StudioClip provides
precise text--motion pairs to the fine tier.

\begin{table}[!ht]
  \centering\footnotesize
  \caption{\textbf{Corpora.} \emph{Source} is the acquisition route, studio
  capture or in-the-wild video reconstruction, and \emph{Caption} what the
  text describes: a clip, a vision--language model's label, or a multi-minute
  take. StudioTake and StudioClip are \textbf{our own in-house capture}
  (\cref{sec:data:ours}); every other row is a public corpus.
  $^{\dagger}$Hours include left--right mirroring across train and held-out
  splits; BONES already contains mirror twins and is counted once.}
  \label{tab:corpus}
  \setlength{\tabcolsep}{4pt}
  \begin{tabular}{llllr}
    \toprule
    Dataset & Source & Caption & Tier & Hours$^{\dagger}$ \\
    \midrule
    HumanML3D~\citep{guo2022humanml3d} & studio & clip & fine & 52.8 \\
    BABEL~\citep{punnakkal2021babel}   & studio & clip & fine & 77.7 \\
    AIST++~\citep{li2021aist}          & studio & clip & fine & 0.4 \\
    GRAB~\citep{taheri2020grab}        & studio & clip & fine & 7.5 \\
    PerMo~\citep{kim2025permo}         & studio & clip & fine & 17.0 \\
    BONES~\citep{luo2026sonic}         & studio & clip & fine & 152.3 \\
    \textbf{StudioClip} (ours, in-house) & studio & clip & fine & 1.8 \\
    \multicolumn{4}{l}{\emph{fine subtotal}} & \textbf{309.6} \\
    \midrule
    OMOMO~\citep{li2023omomo}          & studio & clip & coarse & 14.5 \\
    CIRCLE~\citep{araujo2023circle}    & studio & clip & coarse & 15.5 \\
    TRUMANS~\citep{jiang2024trumans}   & studio & clip & coarse & 4.6 \\
    SAMP~\citep{hassan2021samp}        & studio & clip & coarse & 0.8 \\
    Embody~\citep{mclean2025embody}    & studio & clip & coarse & 60.8 \\
    RoMo~\citep{zhang2026romo}         & wild & VLM & coarse & 173.0 \\
    MMGV~\citep{fan2025motionmillion}  & wild & VLM & coarse & 219.0 \\
    \textbf{StudioTake} (ours, in-house) & studio & take & coarse & 1185.5 \\
    \multicolumn{4}{l}{\emph{coarse subtotal}} & \textbf{1673.7} \\
    \midrule
    \multicolumn{4}{l}{\textbf{total}} & $\mathbf{1983.3}$ \\
    \bottomrule
  \end{tabular}
\end{table}

\ifmoveoverflow
\subsection{Curation details}
\label{sec:suppl:jitter}

\input{sec/mov/par_canon}
\input{sec/mov/par_takes}
\input{sec/mov/par_gates}
\input{sec/mov/par_jitter}
\fi

\subsection{Caption rules}
\label{sec:suppl:captions}

\input{sec/mov/par_align}
\FloatBarrier
\ifmoveoverflow
\section{Evaluation protocol}
\label{sec:suppl:protocol}

\subsection{Test split and sampling configuration}

\input{sec/mov/par_protocol}
\begin{figure}[!t]
  \centering
  \includegraphics[width=\linewidth,height=2.05in,keepaspectratio]{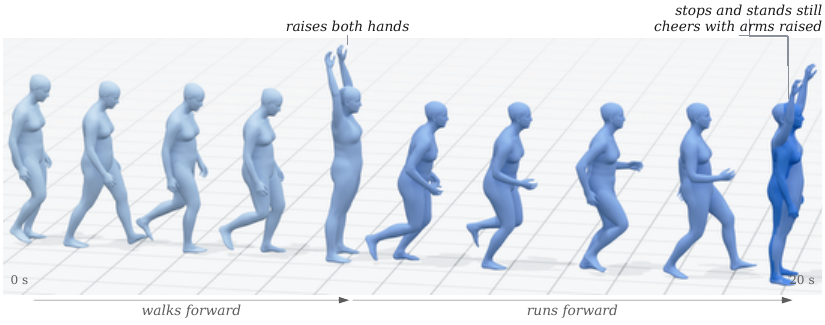}\\[2pt]
  \includegraphics[width=\linewidth]{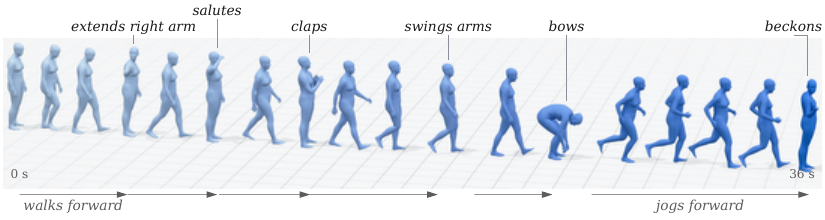}
  \caption{\textbf{Additional long-sequence examples.} Top: five prompts over
  approximately $20$ seconds. Bottom: twelve prompts over $36$ seconds.
  As in \cref{fig:longseq}, poses share a scene and darken with time;
  each displayed pose is aligned to the floor. The lower sequence also
  includes motion postprocessing (\cref{sec:suppl:longseq}).}
  \label{fig:longseq:additional}
\end{figure}

\subsection{Evaluation axes}

\input{sec/mov/par_axes}
\input{sec/mov/par_smooth}
\fi

\section{Additional results}
\label{sec:suppl:qual}

\subsection{Long-sequence examples}
\label{sec:suppl:longseq}

The three examples in \cref{fig:longseq,fig:longseq:additional} correspond
to the accompanying long-sequence videos. Both $36$-second sequences use
the eight-second window configuration. The approximately $20$-second
sequence comes from an earlier checkpoint with a longer sampling window;
it is included as a qualitative example, not in the quantitative comparison.
The second $36$-second sequence (A05) includes heading, floor and contact
corrections, with stationary root motion during in-place segments. For all
three static visualisations, displayed poses are individually placed on the
floor, using one camera and one scale within each scene. These figures show
action composition, not unprocessed root-height accuracy or the absence of
drift. Travelling poses are selected by distance covered, and in-place
actions by representative articulation.

\subsection{Additional qualitative comparisons}

\begin{figure}[t]
  \centering
  \includegraphics[width=\linewidth]{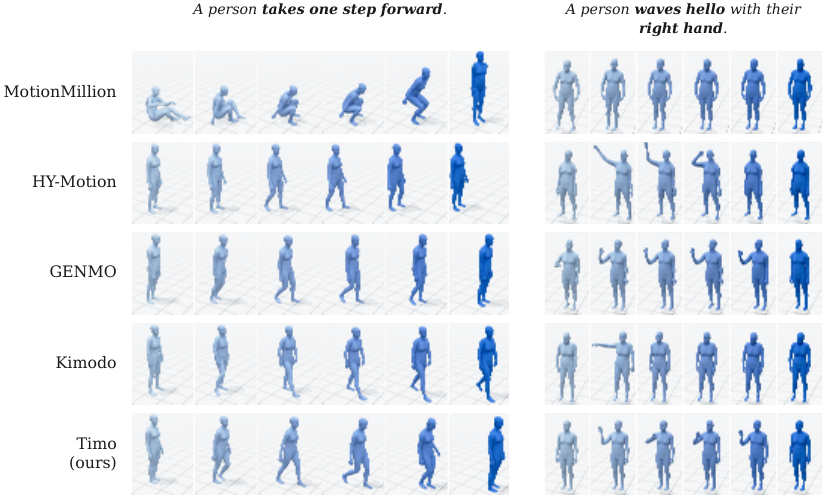}
  \caption{\textbf{Additional qualitative comparisons.} Waving and stepping
  prompts across five methods, with six displayed poses per result and
  a common body scale.}
  \label{fig:suppl:qual-step}
\end{figure}

\cref{fig:suppl:qual-step} extends the comparison in
\cref{fig:qualitative} with waving and stepping prompts. Each column uses
the same camera and body scale across methods. Waving frames cover the
active gesture interval; stepping frames are selected by distance travelled.
The panels illustrate action execution and amplitude without introducing
additional aggregate metrics.

\subsection{Motion coverage}
\begin{figure}[t]
  \centering
  \includegraphics[width=\linewidth,height=0.79\textheight,keepaspectratio]{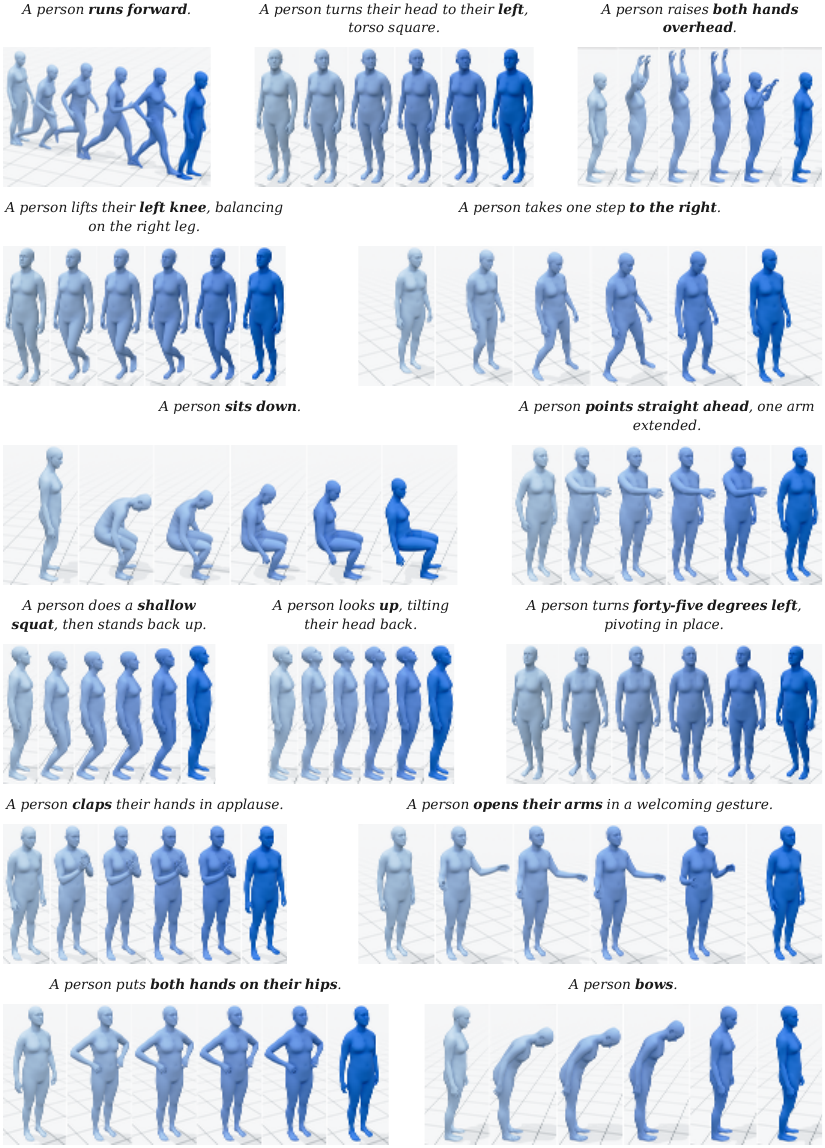}
  \caption{\textbf{\method{} across fourteen prompts}: six moments each, shaded
  light to dark. One panel height is fixed for the whole figure, so the
  $0.5$\,m grid is the same size in every cell and amplitude compares across
  prompts as well as within one.}
  \label{fig:suppl:extra}
\end{figure}

The cross-method comparisons above focus on whether a requested action is
completed. \cref{fig:suppl:extra} complements them with fourteen prompts
from one checkpoint, without per-prompt tuning, covering locomotion, posture
and localised head motion. Mirror pairs are shown once. The shared display
scale reveals movement extent, and the ordered poses show action
development; this is a coverage illustration rather than a benchmark.

\subsection{Physical deployment}
\begin{figure}[t]
  \centering
  \includegraphics[width=\linewidth]{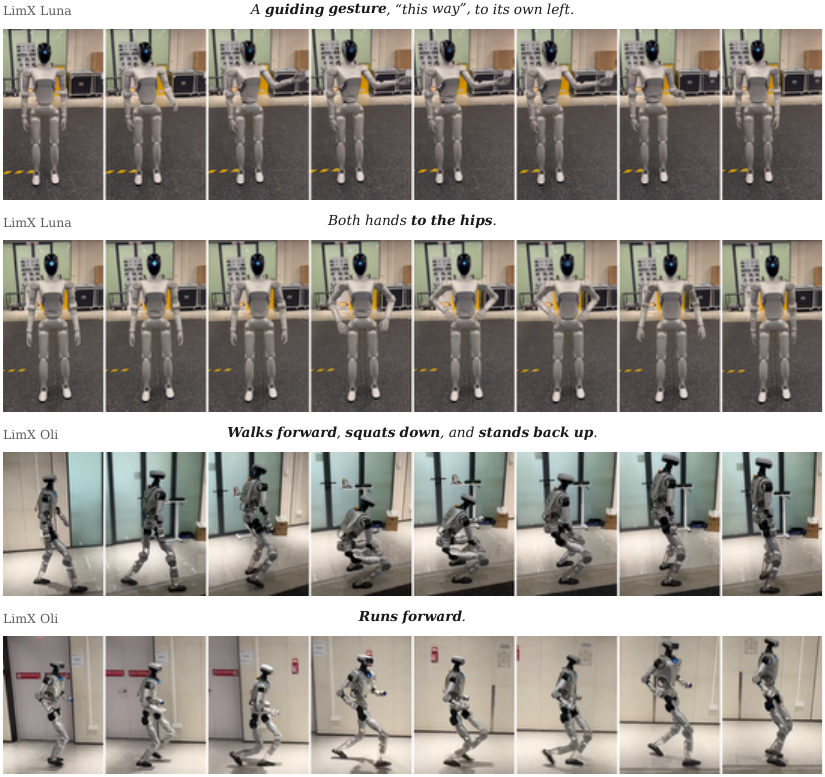}
  \caption{\textbf{Four motions executed on physical humanoids.} Samples from
  \method{}, retargeted to the robot's rig and tracked by the whole-body
  controller of \citet{fu2026cosa05}; eight moments per row. The first two rows
  are a LimX Luna~\citep{limx_luna} and the last two a LimX
  Oli~\citep{limx_oli}. The generator is shared across platforms;
  retargeting and tracking adapt its outputs to each robot. Robot images
  are not shown at a common physical scale.}
  \label{fig:suppl:robot}
\end{figure}

Where the preceding figures show generated human motion,
\cref{fig:suppl:robot} shows its execution after retargeting, expanding the
two-platform overview in \cref{fig:robot}. Four generated motions are
tracked using the interface in \cref{sec:method:robot}; one sequence combines
walking, squatting and standing within a single command. The generator
provides the motion reference, while the robot-specific controller determines
its physical execution. These examples demonstrate that the interface works
on the two platforms, not a success rate over the prompt distribution.

\ifmoveoverflow
\section{Extended related work}
\label{sec:suppl:related}

\input{sec/mov/par_t2m}
\input{sec/mov/par_flowmatch}
\input{sec/mov/par_foundation}
\fi

%% file: sec/mov/par_fuse.tex
For a requested sequence longer than the training window, we maintain one
motion trajectory and extract overlapping crops at each sampling step.
Each crop is processed with its corresponding text conditioning. Predictions
in overlapping regions are blended with linear weights before the next step,
so subsequent updates operate on a shared trajectory. At or below the training
window length, no fusion is applied. The examples in
\cref{sec:exp:deployment} illustrate this procedure; they do not constitute
a quantitative comparison of transition artefacts or long-term drift.

%% file: sec/mov/par_bitident.tex
The procedure is bit-identical to plain sampling at or below $240$ frames.